\documentclass[conference]{IEEEtran}
\IEEEoverridecommandlockouts
\usepackage{cite}
\usepackage{amsmath,amssymb,amsfonts}
\usepackage{algorithmic}
\usepackage{graphicx}
\usepackage{textcomp}
\usepackage{xcolor}
\usepackage{threeparttable}
\usepackage{multirow}
\usepackage{flushend}
\usepackage{hyperref}
\usepackage{bm}
\def\BibTeX{{\rm B\kern-.05em{\sc i\kern-.025em b}\kern-.08em
    T\kern-.1667em\lower.7ex\hbox{E}\kern-.125emX}}
\begin{document}

\title{DD-GCN: Directed Diffusion Graph Convolutional Network for Skeleton-based Human Action Recognition\\
\footnotesize \thanks{ This work is supported by the Key Research and Development Program of China (No. 2022YFC3005401), Key Research and Development Program of China, Yunnan Province (No. 202203AA080009), the Fundamental Research Funds for the Central Universities (No. B230205027), Postgraduate Research \& Practice Innovation Program of Jiangsu Province, the 14th Five-Year Plan for Educational Science of Jiangsu Province (No. D/2021/01/39), the Jiangsu Higher Education Reform Research Project (No. 2021JSJG143) and the 2022 Undergraduate Practice Teaching Reform Research Project of Hohai University. \\ \indent Codes are available at \href{https://github.com/shiyin-lc/DD-GCN}{https://github.com/shiyin-lc/DD-GCN}\\ \indent \textsuperscript{*} Corresponding author.}
}

\author{
	\IEEEauthorblockN{Chang Li$^{1,2}$, Qian Huang$^{1,2}$\textsuperscript{*}, Yingchi Mao$^{1,2}$\textsuperscript{*}}
	\IEEEauthorblockA{$^1$ Key Laboratory of Water Big Data Technology of Ministry of Water Resources, Hohai University, Nanjing, China}
	\IEEEauthorblockA{$^2$ School of Computer and Information, Hohai University, Nanjing, China}
	\IEEEauthorblockA{\{lichang, huangqian, yingchimao\}@hhu.edu.cn}
}
\maketitle

\begin{abstract}
Graph Convolutional Networks (GCNs) have been widely used in skeleton-based human action recognition. In GCN-based methods, the spatio-temporal graph is fundamental for capturing motion patterns. However, existing approaches ignore the physical dependency and synchronized spatio-temporal correlations between joints, which limits the representation capability of GCNs. To solve these problems, we construct the directed diffusion graph for action modeling and introduce the activity partition strategy to optimize the weight sharing mechanism of graph convolution kernels. In addition, we present the spatio-temporal synchronization encoder to embed synchronized spatio-temporal semantics. Finally, we propose Directed Diffusion Graph Convolutional Network (DD-GCN) for action recognition, and the experiments on three public datasets: NTU-RGB+D, NTU-RGB+D 120, and NW-UCLA, demonstrate the state-of-the-art performance of our method.
\end{abstract}

\begin{IEEEkeywords}
Skeleton-based action recognition, graph convolutional network, spatial temporal modeling
\end{IEEEkeywords}

\section{Introduction}
\label{sec:intro}
\par Human action recognition is a hot topic with broad applications, including video surveillance, human-computer interaction, and abnormal behavior monitoring. Human actions can be described by multimodal data, such as RGB videos, depth videos, and skeleton sequences etc. Due to its compactness and robustness to environmental variations, skeleton-based methods have attracted increasing attention. In essence, human skeleton sequences can be represented as isomorphic graphs. Taking advantage of this structure, Graph Convolutional Networks (GCNs) have achieved promising results in human action recognition. 
\begin{figure}[t]
	\centerline{\includegraphics[width=0.45\textwidth,height=0.33\textwidth]{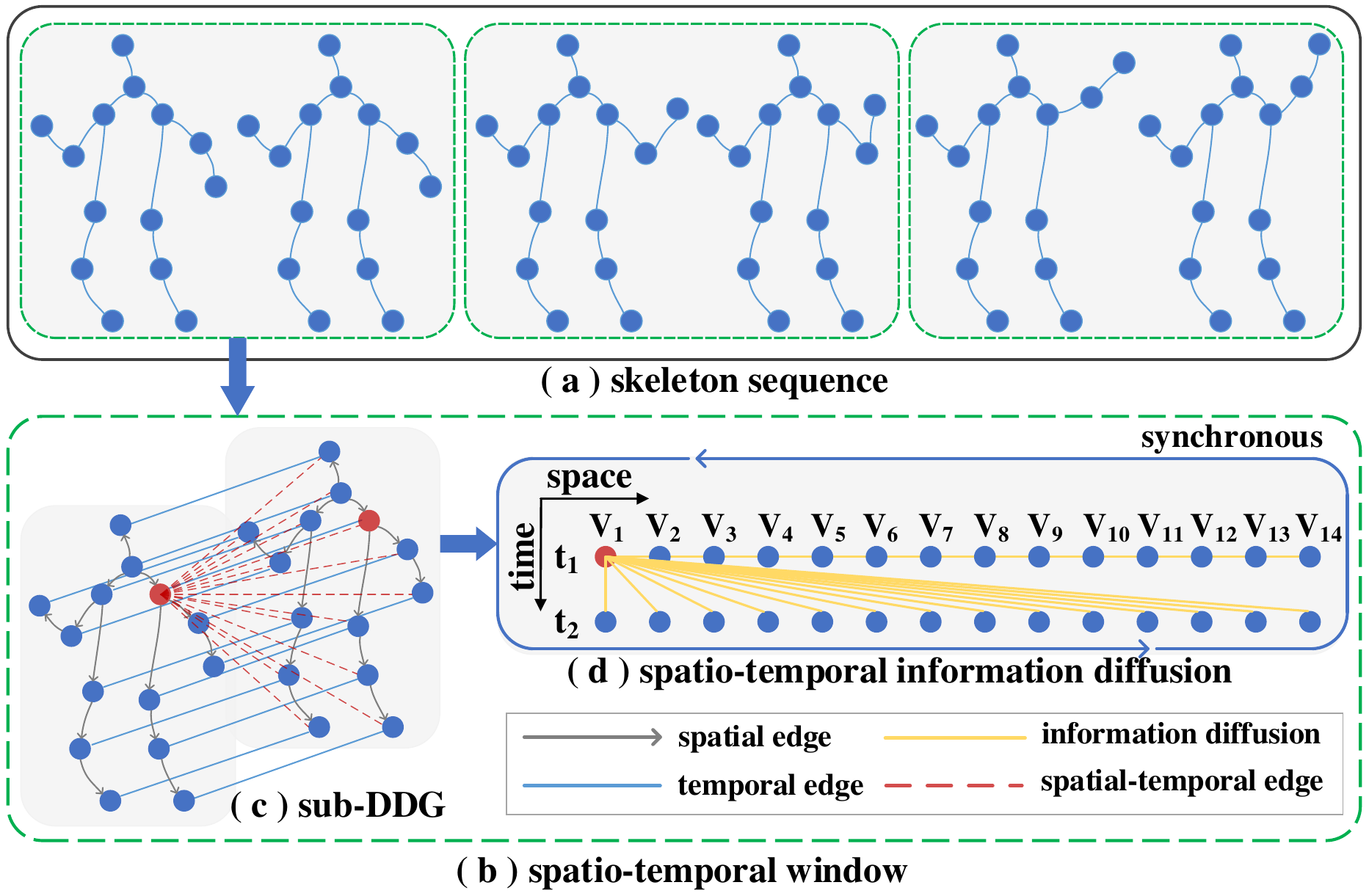}}
	\caption{Illustration of Directed Diffusion Graph(DDG). The spatio-temporal window (b) divides the skeleton sequence (a) into multiple spatio-temporal neighborhood sets, each of which is modeled as sub-DDG(c) to synchronize inter-frame and intra-frame information(d).}
	\label{fig:1}
\end{figure}
\par One of the fundamental issues in GCN-based methods is action modeling. Yan et al. \cite{RN11} first constructed the spatio-temporal graph to describe skeleton sequence. They considered the bones as spatial edges and connected the same joints in consecutive frames as temporal edges. Generally, the methods \cite{RN11, RN14, RN17} based on this vanilla spatio-temporal graph have two shortcomings. First, these methods neglected the physical dependency between joints. For example, the wrist moves with the forearm around the elbow. Second, the temporal edges only bind the identical joints in adjacent frames, which limits the spatio-temporal information to diffuse between joints. It is suboptimal because human actions involve highly concurrent space-time interactions of joints, not just the separate trajectory of each joint. For example, when walking, the left arm and right arm tend to move oppositely, while the left arm and right leg have the same tendency. This demonstrates that each joint diffuses the information in localized spatio-temporal joint sets rather than only within itself. To address these challenges, each frame is described as the Directed Diffusion Graph (DDG). The directed edges between nodes not only reflect the connections but also indicate their relative movement modes. Furthermore, we focus on boosting inter-frame interactions by connecting multiple joints in long-range frames, which benefits extracting intra-frame and inter-frame features of all the joints involved in human actions, as shown in Fig. 1. 
\par Another critical task is action representation. Most existing methods first perform Spatial Graph Convolution (SGC) to extract spatial features which are then fed into a temporal convolution module to capture spatio-temporal patterns. However, these methods suffer from two drawbacks. Firstly, they always utilize spatial configuration partitioning strategy coined in ST-GCN\cite{RN11} to learn the weights for various neighborhoods. This strategy disregards the physical dependency between joints and restricts the representation capability of GCNs. Secondly, these methods \cite{RN14, RN13,  RN15, RN16} extract spatial configuration and temporal dynamics from separate components. This factorized formulation breaks the synchronized spatio-temporal relations between joints. To address these limitations, we propose the activity partition strategy according to the hinged structure of skeletons to optimize graph convolution kernels. Moreover, we introduce a Spatio-temporal Synchronous Encoder (STSE) to embed spatio-temporal action patterns simultaneously, thus facilitating more discriminative features.
\par By coupling the above proposals, we present the Directed Diffusion Graph Convolutional Network (DD-GCN) for skeleton-based human action recognition. Our approach achieves state-of-the-art performance on three datasets: NTU-RGB+D, NTU-RGB+D 120, and NW-UCLA. The main contributions of our work are as follows:
\begin{itemize}
	\item We construct DDG to model actions, which preserves the kinematic dependence of skeletons and synchronization of spatio-temporal information diffusion.
	\item We propose the activity partition strategy to optimize the weight allocation of SGC, which facilitates discriminative features and is universal in GCNs.
	\item We design the STSE module to capture spatial and temporal motion patterns synchronously, which can be embedded in other GCNs to boost performance.
	\item Based on these proposals, we present a novel learning framework named DD-GCN, and the extensive experiments on three public datasets have proved its superiority. 
\end{itemize}
\section{Related Work}
\textbf{Spatio-temporal graph for action modeling}. Conventional methods always model the skeleton data as a sequence of vectors or a pseudo-image to be processed by RNNs or CNNs. However, these approaches ignore the physical structure between joints and bones. Yan et al. \cite{RN11} first constructed the spatial-temporal graph for action modeling, where the joints within one frame are connected with spatial edges according to skeleton structure, and temporal edges connect the same joint in the consecutive frame. This static topology limits GCNs to capture long-range dependencies between joints. Shi et al. \cite{RN14} proposed an adaptive graph convolutional network to learn the spatial topology in an end-to-end manner adaptively. But undirected connections fail to reflect the complex correlations between joints adequately. To refine the topology, Shi et al. \cite{RN12} represented the skeleton data as a directed acyclic graph, where the vertex closer to the predefined center vertex points to the farther one. However, the kinematic dependency that one joint moves around another is ignored, and the methods above all focus on spatial graph, but the optimization of the inter-frames temporal edges is rarely considered. 
\par \textbf{Partition strategy of graph convolution}. There is no rigid arrangement naturally exists in graphs. Thus, it is difficult to establish the correspondence between various neighbors and weights. To solve this problem, Yan et al. \cite{RN11} propose three partition strategies to group neighbors and the neighbors in the identified subset sharing unique weight. For the uni-labeling strategy, all the neighbors are indistinctive. That is to say, the feature vectors of every neighboring node will have an inner product with the same weight vector. Unlike this, the distance partition strategy divides neighbors based on their distance from the root node. The spatial configuration partitioning strategy is widely used, which separates neighborhoods into three subsets: (1) the root node itself, (2) the centripetal subset, which is closer to the body barycenter than the root, otherwise (3) the centrifugal subset. These strategies are not optimal because they limit the full use of the intrinsic characteristic of joints. However, there has been a paucity of research on graph convolution partition strategies.
\par \textbf{Attention mechanism in GCN-based methods}. Chen et al. \cite{RN17} employed channel attention in GCN to compute the correlation between joints and then dynamically refined the topology. Qian et al. \cite{RN19} introduced a symmetry trajectory attention module to measure the relation between the left and right body and then imposed a part relation attention module to explore the relationships of each part. Zhu et al. \cite{RN16} applied relative attention to model the interaction between interacted skeletons and then constructed dynamic relational graphs. Qiu et al. \cite{RN18} devised a two-branch network that utilized graph convolution units and the temporal attention unit to model long-range dependency and extract local and global spatio-temporal features. Bai et al. \cite{RN20} decomposed spatio-temporal features into spatial and temporal ones, and global and temporal attention modules were performed, respectively. Unlike these methods, we focus on the synchronized spatio-temporal information diffused between inter-frames.
\begin{figure*}[t]
\centerline{\includegraphics[width=0.9\textwidth,height=0.45\textwidth]{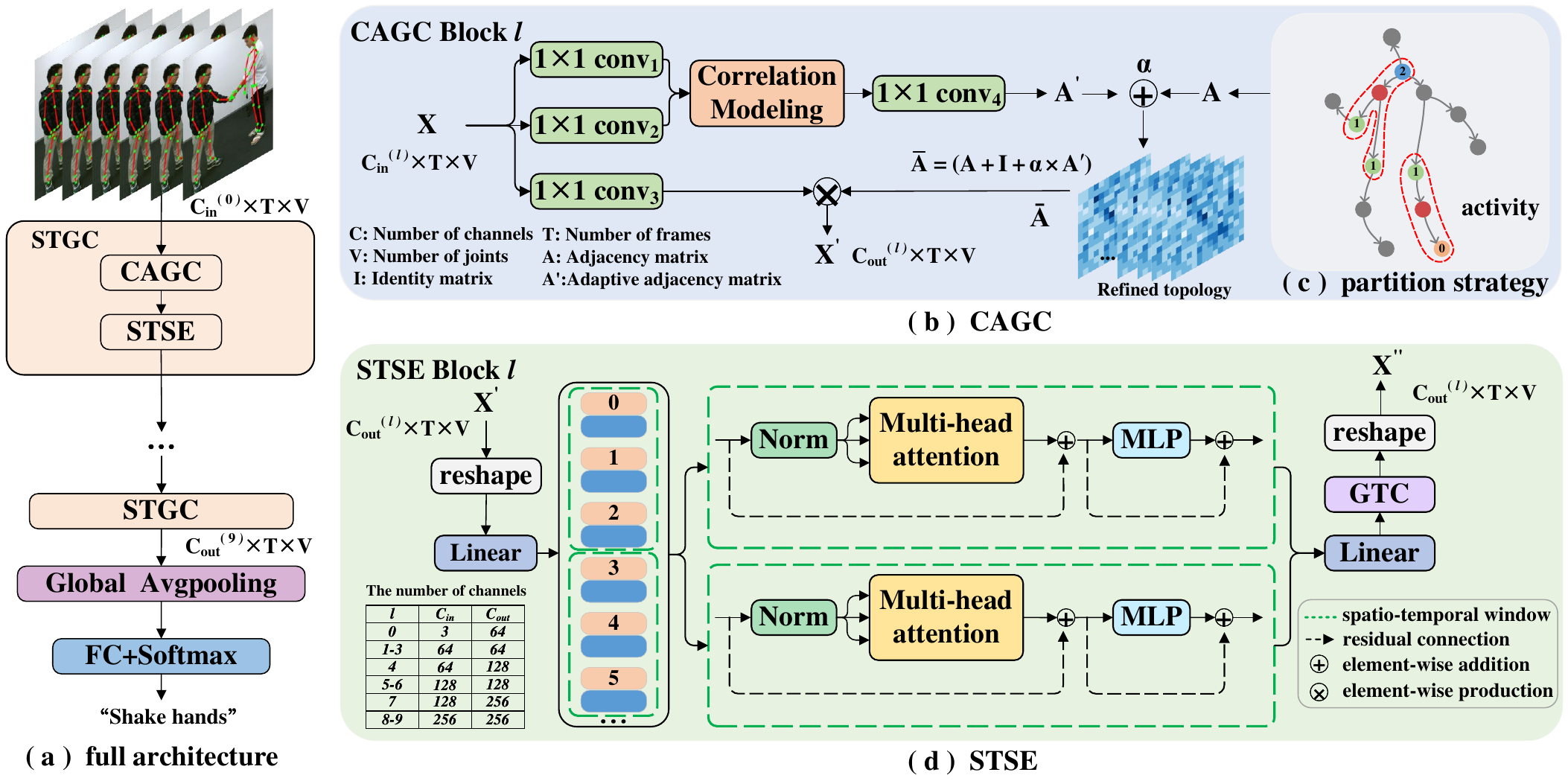}}
\caption{The overview of DD-GCN. (a) It has ten STGC layers containing two modules: CAGC (b) and STSE(d). After global average pooling, Softmax is utilized for action classification. CAGC is the unit of channel-wise correlation modeling and graph convolution with activity partition strategy(c). STSE employs MSA and GTC for synchronized spatio-temporal embedding.}
\label{fig:2}
\end{figure*}
\section{Method}
\subsection{Pipeline Overview}
The overview of the proposed DD-GCN is illustrated in Fig. 2. The input of DD-GCN is skeleton sequences denoted as $X \in \mathbb{R}^{T \times V \times C}$, where $T$, $V$, and $C$ denotes the number of frames, joints, and channels, respectively. DD-GCN has ten Spatial Temporal Graph Convolution (STGC) layers for hierarchical motion representation. Each layer contains two modules, i.e., Channel-wise Adaptive Graph Convolution (CAGC) module and STSE. CAGC first constructs DDG for action modeling. On this basis, it obtains channel-wise refined adjacency matrix by correlation modeling function \cite{RN17} and adopts activity partition strategy for adaptive joints embedding. Then, the rearranged joint sequences are put into STSE, where the normalized feature in parallel spatio-temporal windows is aggregated synchronously by Multi-head Self-attention (MSA) to reserve diffused spatio-temporal signals. Afterwards, the Group Temporal Convolution (GTC) calculates the window-wise aggregated feature and passes them to the next layer. Finally, the output feature will send to the Softmax classifier for action recognition. Besides, we boost performance by ensemble joint and bone stream, considering the complementarity of multimodal data.
\subsection{Spatio-temporal Directed Diffusion Graph}
Human actions can be described by skeleton sequences and modeled by spatio-temporal graphs. Considering the kinematic dependency and high concurrency of the joints, we construct the DDG, denoted as $G$, for action modeling as shown in Fig. 1. Each skeleton is represented as a directed graph, and the directed edge indicates the dependency between joints, e.g., $v_i \rightarrow v_j$ manifests that $v_j$ moves around $v_i$. Because interactions between joints are spatio-temporal synchronized and localized, the temporal edges are extended to multiple joints within spatio-temporal windows for inter-frame information diffusion. By adjusting the window, actions can be described by multiple sub-DDGs represented as $g_i$, i.e., $G=\left\{g_1, g_2, \ldots, g_n\right\}$. The spatial edges are reflected by the adjacency matrix $A \in \mathbb{R}^{V \times V}$, and the spatial neighborhood of $v_i$ can be defined as $\mathcal{N}_{v_i}^S=\left\{v_j \mid A_{i, j} \neq 0\right\}$.  The localized spatio-temporal neighborhood of $v_i$ is defined as $\mathcal{N}_{v_i}^{S T}=\left\{v_j \mid v_i, v_j \in g\right\}$.
\subsection{Channel-wise Adaptive Graph Convolution}
Since there is no rigid arrangement naturally exists in graphs, it is difficult to establish the correspondence between neighbor vertices and weights. To solve this problem, Yan et al. \cite{RN11} divided vertexes into $K$ subsets, and mapped each vertex with a unique weight according to subset index. On this basis, the spatial graph convolution operation on $v_i$  is formulated as:
\begin{equation}
f_{\text {out }}\left(v_i\right)=\sum_{v_j \in \mathcal{N}_{v_i}} \frac{1}{Z_{i, j}} f_{\text {in }}\left(\mathcal{P}\left(v_i, v_j\right)\right) \cdot W\left(\mathcal{M}\left(v_i, v_j\right)\right)
\end{equation}
where $Z_{i,j}$ denotes the cardinality of the corresponding subset $\mathcal N_{v_i}^k$. It balances the contribution of each subset. $f_{in}$ is the input feature map $X \in \mathbb{R}^{T \times V \times C}$. $\mathcal{P}$ is the partition strategy determining $v_j$  belongs to which subset, and $\mathcal{M}$ is the mapping function that is the subset index of $v_j$. $W$ is the weight vector with $K$ dimension.
\par \textbf{Activity partition strategy}. The partitioning strategy not only determines the number of subsets, i.e., the convolutional kernel size, but also constrains parameter sharing. Therefore, the subset partitioning strategy is essential for SGC. For actions, the contribution of joints depends on their own activity. Hands and feet are critical parts of human movements, and they are leaf nodes. Inspired by this, we employ out degree as a measure and propose the activity partition strategy to optimize the graph convolution operation, as illustrated in Fig. 2(c). Formally,
\begin{equation}
\mathcal{M}\left(v_i, v_j\right)=\left\{\begin{array}{l}
0, \text { if } \mathcal{D}\left(v_j\right)=0 \\
1, \text { if } \mathcal{D}\left(v_j\right)=1 \\
2, \text { if } \mathcal{D}\left(v_j\right) \geq 2
\end{array}\right.
\end{equation}
where $\mathcal{D}$  is the out-degree function. Besides, we adopt channel-wise correlation modeling \cite{RN17} for adaptive topologies as depicted in Fig. 2(b). Specifically, CAGC can be formulated by  
\begin{equation}
f_{\text {out }}=\sigma\left(\sum_k^K f_{\text {in }}\left(D_k^{-\frac{1}{2}} \bar{A}_k D_k^{-\frac{1}{2}}\right) W_k\right)
\end{equation}
where $\bar{A}=\left(A+I+\alpha \cdot A^{\prime}\right)$. $I$ is the identity matrix which means self-connection. $A^{\prime}\in \mathbb{R}^{V \times V \times C}$ is the channel specific correlations and $\alpha$ is a trainable scalar. $\bar{A}_k$ is the adjacency matrix of $\mathcal{N}_{v_i}^k$. $D_k$ is the diagonal degree matrix of $\bar{A}_k$. $W_k$ is the weight vector and $k=\mathcal{M}\left(v_i,v_j \right)$. Leveraging activity partition strategy, the CAGC can reasonably assign weights and extract action-specific features, contributing to the strong representation capability of DD-GCN.
\subsection{Spatio-temporal Synchronous Encoder}
In the view that human actions contain highly concurrent spatio-temporal dynamics of joints, we propose STSE to capture synchronized spatio-temporal motion patterns. The correlation between joints is local due to the skeleton structure and time step. For example, the action consists of multiple sub-actions, and joints consisting of the same sub-action are highly correlated. To reduce computing complexity and redundant relations, we split the skeleton sequence by non-overlapping windows as shown in Fig. 2(d). 
\par In particular, given the window size $M \times N$, STSE splits $X \in \mathbb{R}^{T \times V \times C}$  into $n$ windows, where $n= T/M\times V/N$, and $X=\left\{X_1, X_2, \ldots, X_n\right\}$. The localized joints are arranged as a sequence according to sub-DDG, and joints can be regarded as tokens in transformer\cite{RN1}. Then MSA is applied to obtain synchronized spatio-temporal relationships. The linear embedded joint sequences are denoted as $Q$, $K$, $V$, and then the scaled dot-production is conducted to compute similarity. For each head, the attention can be formulated as follows
\begin{equation}
\text { Attention }(\mathrm{Q}, \mathrm{K}, \mathrm{V})=\operatorname{Softmax}\left(\frac{Q K^T}{\sqrt{C^{\prime}}}+B\right) \mathrm{V}
\end{equation}
where $B$ is the relative position bias which is introduced to retain primary structures. Every head is concatenated and then projected for the feature of $X_i$. The complete motion embedding is obtained by the following equation.
\begin{equation}
\mathrm{X}=\text { Concat }\left(X_1, X_2, \ldots, X_n\right)
\end{equation}
MSA allows the model to extract discriminative features from different representation subspaces. Considering the continuity of spatio-temporal information, we employ GTC to implement window-wise feature aggregation. 
\begin{equation}
X_{STS}=\operatorname{Conv} 2 D[\Gamma \times 1](X)
\end{equation}
where $\Gamma \times 1$ is the kernel size. In addition, shortcut and layer normalization are designed for stability. STSE exposes the essential spatio-temporal information flow of actions and thus improves the performance of DD-GCN. 
\section{Experiments and discussions}
\subsection{Datasets}
\par \textbf{NTU RGB+D.} NTU RGB+D  contains 56,880 samples which are categorized into 60 actions. These samples are performed by 40 volunteers and captured by three Microsoft Kinect v2 cameras from various views concurrently. There are two experiments settings: cross-subject and cross-view\cite{RN8}. In cross-subject setup, training set comes from half of the subjects, and testing set comes from the other. In cross-view setup, training set comes from camera ID 2 and 3, and the testing set comes from camera ID 1.
\par \textbf{NTU RGD+D 120.} NTU RGD+D 120 is extended from NTU RGB+D with 60 extra classes. It has 114,480 samples conducted by 106 subjects. The recommended benchmarks are cross-subject (as mentioned above) and cross-setup\cite{RN9}. Training data comes from samples with even setup IDs, and testing data comes from samples with odd setup IDs in cross-setup setting. 
\par \textbf{NW-UCLA.} NW-UCLA\cite{RN10}  is captured by three Kinect cameras. It has 1494 video clips and covers ten classes. Each action is performed by 10 subjects. The data from first two cameras are training set and the others are testing set.
\subsection{Training Details}
We utilize the PyTorch deep learning framework to develop our models and optimize the training with Adam optimizer. Our models are trained for 80 epochs with batch size 64. The learning rate is set to 0.1, which is ten times smaller every ten steps. We adopt the data preprocessing following \cite{RN17} on three datasets.
\subsection{Ablation Study}
\par The ablation experiments are conducted on NTU RGB+D dataset with the cross-subject setup, and we only input bone data as default. We discuss the influence of parameter settings and the effectiveness of the proposed components.
\begin{table}[b]
	\renewcommand\arraystretch{1.1}
	\begin{center}
		\caption{Comparisons of models with various parameters.} \label{tab:1}
		\begin{tabular}{|l|c|c|l|c|c|}
			\hline
			\textbf{$\bm{Win\_ size}$} & \textbf{$\bm{h}$} & \textbf{Acc. (\%)}& \textbf{$\bm{Win\_ size}$}& \textbf{$\bm{h}$} &  \textbf{Acc. (\%)}\\
			\hline
			$4 \times 5$ & 4 & $89.49$ &$4 \times 25$ & 2 & $88.52$ \\
			$8 \times 25$ & 4 & $90.08$ & $4 \times 25$ & 8 & $88.34$\\
			$32 \times 25$ & 4 & $88.60$ & $4 \times 25$ & 4 & $\mathbf{90.52}$ \\			
			\hline
		\end{tabular}
	\end{center}
\end{table}
\par\textbf{Parameter selection}. The experimental results in Table \uppercase\expandafter{\romannumeral1} shows that the window size denoted as ${Win\_ size}$ influences spatio-temporal information's diffusion intensity. If the window is too large, it will lead to redundant correlation and computational cost. In addition, the unit of sub-DDG should be sub-action as the accuracy will be impaired when the window's width is smaller than $V$, i.e., the minimum unit of sub-DDG should be a complete skeleton. The possible explanation is that human motion modeling relies on the integrity of the body structure. Besides, the number of heads denoted as $h$ is also a vital factor. Table \uppercase\expandafter{\romannumeral 1} suggests that more heads may not be optimal. Verified by the experiments,, we set $Win\_ size$ to $4 \times 25$, and $h$ is set to 4.
\par\textbf{Comparisons of partition strategies}. We validate the proposed activity partition strategy by comparing it against other strategies in \cite{RN11}. In addition, Table \uppercase\expandafter{\romannumeral 2} demonstrated that our strategy is compatible with other GCNs. The experimental results in Table \uppercase\expandafter{\romannumeral 2} verify its superiority. Based on activity partition strategy, we developed the CAGC module. As presented in Table \uppercase\expandafter{\romannumeral 2}, our proposed DD-GCN, which incorporates CAGC, surpasses the orignal ST-GCN and AGCN by 6.87\% and 1.68\%, respectively.
\par \textbf{The effectiveness of STSE module}. We validate the effectiveness of the STSE from two aspects as reported in Table \uppercase\expandafter{\romannumeral 3}. First, the accuracy of our method is dropped by 0.67\% after the STSE module is dismantled.  Second, the ST-GCN and AGCN combined with the STSE module improved their performance by 4.2\% and 0.69\%, respectively. It further explains that the STSE module is practical and plug-and-play. In addition, the relative position bias is valuable, which increases the accuracy by 0.25\%. 
\begin{table}
	\renewcommand\arraystretch{1.15}
	\begin{center}
		\begin{threeparttable}[t]
			\caption{Comparisons with different partition strategies.} \label{tab:2}
			\begin{tabular}{|l|c|c|c|c|}
				\hline
				\textbf{Methods} & \textbf{Uniform} & \textbf{Distance}  & \textbf{Spatial}  & \textbf{Activity} 
				\\
				\hline
				ST-GCN\cite{RN11} $^*$ & $80.84$ & $83.33$ & $83.65$ & $\mathbf{84.89}$ \\
				AGCN\cite{RN14} $^*$ & $88.29$ & $88.51$ & $88.84$ & $\mathbf{89.00}$ \\
				DD-GCN & $89.71$ & $90.04$ & $90.13$ & $\mathbf{90.52}$ \\
				\hline
			\end{tabular}
			\begin{tablenotes}\footnotesize
				\item Those marked with $*$ are the methods we reproduced.
			\end{tablenotes}
		\end{threeparttable}
	\end{center}
\end{table}
\begin{table}[t]
	\renewcommand\arraystretch{1.15}
	\begin{center}
		\begin{threeparttable}[c]
			\caption{Comparisons of models with/without STSE and relative position bias.} \label{tab:3}
			\setlength{\tabcolsep}{6mm}
			\begin{tabular}{|l|c|}
				\hline
				\textbf{Backbone}  & \textbf{Accuracy (\%)}
				\\
				\hline
				ST-GCN\cite{RN11} $^*$( w/ STSE) & $83.65(87.85) \uparrow \mathbf{4.20}$ \\
				AGCN\cite{RN14} $^*$(w/ STSE) & $88.84(89.53) \uparrow \mathbf{0.69}$ \\
				DD-GCN(w/o STSE) & $90.52(89.85) \downarrow \mathbf{0.67}$ \\
				DD-GCN $($ w/o PE) & $90.52(90.27) \downarrow \mathbf{0.25}$ \\
				\hline
			\end{tabular}
			\begin{tablenotes}\footnotesize
				\item {Those marked with $*$ are the methods we reproduced.}
			\end{tablenotes}
		\end{threeparttable}
	\end{center}
\end{table}
\begin{table}[b]
	\renewcommand\arraystretch{1.1}
	\begin{center}
		\caption{Comparisons on NTU RGB+D dataset.} \label{tab:4}
		\begin{tabular}{|l|c|c|}
			\hline 
			\multirow{2}{*}{\textbf{Methods}} & \multicolumn{2}{|c|}{\textbf{Accuracy (\%)}}\\
			& \multicolumn{1}{|l}{ \textbf{Cross-subject} } & \textbf{Cross-view} \\
			\hline 
			CTR-GCN\cite{RN17} & $92.4$ & $96.8$ \\
			Graph2Net\cite{RN22} & $90.1$ & $96.0$ \\
			GLTA-GCN\cite{RN18} & $61.2$ & $81.2$ \\
			MKE-GCN\cite{RN23} & $91.8$ & $96.2$ \\
			H-GCN(fusion)\cite{RN20} & $91.6$ & $96.2$ \\
			SMotif-GCN\cite{RN27} & $90.5$ & $96.1$ \\
			SA-GCN\cite{RN19} & $92.6$ & $96.7$ \\
			\hline
			AimCLR\cite{RN25} & $86.9$ & $92.8$ \\
			Ta-CNN++\cite{RN26} & $90.7$ & $95.1$ \\
			\hline
			DD-GCN (bone) & $90.5$ & $95.7$ \\
			DD-GCN (fusion) & $\mathbf{92.6}$ & $\mathbf{96.9}$ \\
			\hline
		\end{tabular}
	\end{center}
\end{table}
\begin{table}[t]
	\renewcommand\arraystretch{1.1}
	\begin{center}
		\caption{Comparisons on NTU RGB+D 120 dataset.} \label{tab:5}
		\begin{tabular}{|l|c|c|}
			\hline 
			\multirow{2}{*}{\textbf{Methods} } & \multicolumn{2}{|c|}{\textbf{Accuracy (\%)}} \\
			& \multicolumn{1}{|l}{ \textbf{Cross-subject} } & \textbf{Cross-setup}  \\
			\hline 
			Graph2Net\cite{RN22} & $86.0$ & $87.6$ \\
			GLTA-GCN\cite{RN18} & $49.1$ & $51.1$ \\
			H-GCN\cite{RN20} & $88.9$ & $90.0$ \\
			SMotif-GCN\cite{RN27}& $87.1$ & $87.7$ \\
			CTR-GCN(bone)\cite{RN17} & $85.7$ & $87.5$ \\
			CTR-GCN(fusion)\cite{RN17} & $88.7$ & $90.1$ \\
			\hline
			Ta-CNN++\cite{RN26} & $85.7$ & $87.3$ \\
			AimCLR\cite{RN25} & $80.1$ & $80.9$ \\
			\hline
			DD-GCN (bone) & $86.1$ & $87.6$ \\
			DD-GCN (fusion) & $\mathbf{88.9}$ & $\mathbf{90.2}$ \\
			\hline
		\end{tabular}
	\end{center}
\end{table}
\begin{table}[h]
	\renewcommand\arraystretch{1.1}
	\begin{center}
		\caption{Comparisons on NW-UCLA dataset.} \label{tab:cap}
		\setlength{\tabcolsep}{6mm}
		\begin{tabular}{|l|c|}
			\hline
			\textbf{Methods} & \textbf{Accuracy (\%)}\\
			\hline
			DC-GCN+ADG\cite{RN29} & $95.3$ \\
			ShiftGCN++\cite{RN21} & $95.0$ \\
			GCN-HCRF\cite{RN31} & $91.5$ \\
			Graph2Net\cite{RN22} & $95.3$\\
			CTR-GCN(fusion)\cite{RN19}& $96.5$ \\
			FGCN\cite{RN30} & $95.3$ \\
			\hline
			DD-GCN(fusion) & $\mathbf{96.7}$ \\
			\hline
		\end{tabular}
	\end{center}
\end{table}
\subsection{Comparison with the State-of-the-art}
We compare our model with the state-of-the-art methods on three public datasets: NTU RGB+D, NTU RGB+D 120, and NW-UCLA. The experiment results are reported in Table \uppercase\expandafter{\romannumeral 4}-\uppercase\expandafter{\romannumeral 6}. Fusion means the two-stream network with joint and bone data. For the NTU RGB+D and NTU RGB+D 120 datasets, DD-GCN outperforms the state-of-the-art models, especially those without graph convolutions \cite{RN25, RN26}. In addition, DD-GCN achieves the highest performance on the NW-UCLA dataset as well, which indicates that the proposed DDG can effectively model daily behaviors and complex interactive actions. The extensive experiments fully demonstrate the superiority of DD-GCN and its components: CAGC with activity partition strategy and STSE.
\section{Conclusion}
We propose the Directed Diffusion Graph Convolutional Network (DD-GCN) for skeleton-based human action recognition.  We construct the directed diffusion graph to model the kinematic dependencies of intra-frame joints and enhance temporal correlations between inter-frame joints. Then, we devise the activity partition strategy to optimize the weight allocation of the graph convolution. Besides, we design the spatio-temporal synchronous encoder to embed diffused spatio-temporal information simultaneously. The experiments on three datasets have proved the superiority of DD-GCN.
\bibliographystyle{IEEEbib}
\bibliography{ICMEref}

\begin{thebibliography}{10}

\bibitem{RN11}
Sijie Yan, Yuanjun Xiong, and Dahua Lin,
\newblock ``Spatial temporal graph convolutional networks for skeleton-based
  action recognition,''
\newblock in {\em AAAI Conference on Artificial Intelligence}, 2018, pp.
  7444--7452.

\bibitem{RN14}
Lei Shi, Yifan Zhang, Jian Cheng, and Hanqing Lu,
\newblock ``Two-stream adaptive graph convolutional networks for skeleton-based
  action recognition,''
\newblock in {\em Computer Vision and Pattern Recognition (CVPR)}, 2019, pp.
  12018--12027.

\bibitem{RN17}
Yuxin Chen, Ziqi Zhang, Chunfeng Yuan, Bing Li, Ying Deng, and Weiming Hu,
\newblock ``Channel-wise topology refinement graph convolution for
  skeleton-based action recognition,''
\newblock in {\em International Conference on Computer Vision (ICCV)}, 2021,
  pp. 13339--13348.

\bibitem{RN13}
Matthew Korban and Xin Li,
\newblock ``Ddgcn: A dynamic directed graph convolutional network for action
  recognition,''
\newblock in {\em European Conference on Computer Vision}, 2020, pp. 761--776.

\bibitem{RN15}
Ziyu Liu, Hongwen Zhang, Zhenghao Chen, Zhiyong Wang, and Wanli Ouyang,
\newblock ``Disentangling and unifying graph convolutions for skeleton-based
  action recognition,''
\newblock in {\em Computer Vision and Pattern Recognition (CVPR)}, 2020, pp.
  140--149.

\bibitem{RN16}
Liping Zhu, Bohua Wan, Chengyang Li, Gangyi Tian, Yi~Hou, and Kun Yuan,
\newblock ``Dyadic relational graph convolutional networks for skeleton-based
  human interaction recognition,''
\newblock {\em Pattern Recognition}, vol. 115, pp. 107920, 2021.

\bibitem{RN12}
Lei Shi, Yifan Zhang, Jian Cheng, and Hanqing Lu,
\newblock ``Skeleton-based action recognition with directed graph neural
  networks,''
\newblock in {\em Computer Vision and Pattern Recognition (CVPR)}, 2019.

\bibitem{RN19}
Ruihao Qian, Jiewen Wang, Jianxiu Wang, and Shuang Liang,
\newblock ``Structural attention for channel-wise adaptive graph convolution in
  skeleton-based action recognition,''
\newblock in {\em International Conference on Multimedia and Expo (ICME)},
  2022, pp. 01--06.

\bibitem{RN18}
Haoyue Qiu, Yuan Wu, MengMeng Duan, and Cheng Jin,
\newblock ``Glta-gcn: Global-local temporal attention graph convolutional
  network for unsupervised skeleton-based action recognition,''
\newblock in {\em International Conference on Multimedia and Expo (ICME)},
  2022, pp. 1--6.

\bibitem{RN20}
Ruwen Bai, Min Li, Bo~Meng, Fengfa Li, Miao Jiang, Junxing Ren, and Degang Sun,
\newblock ``Hierarchical graph convolutional skeleton transformer for action
  recognition,''
\newblock in {\em International Conference on Multimedia and Expo (ICME)},
  2022, pp. 01--06.

\bibitem{RN1}
Ashish Vaswani, Noam Shazeer, Niki Parmar, Jakob Uszkoreit, et~al.,
\newblock ``Attention is all you need,''
\newblock in {\em Neural Information Processing Systems}, 2017, p. 6000–6010.

\bibitem{RN8}
Amir Shahroudy, Jun Liu, Tian-Tsong Ng, and Gang Wang,
\newblock ``Ntu rgb+d: A large scale dataset for 3d human activity analysis,''
\newblock in {\em Computer Vision and Pattern Recognition (CVPR)}, 2016, pp.
  1010--1019.

\bibitem{RN9}
Jun Liu, Amir Shahroudy, Mauricio Perez, Gang Wang, et~al.,
\newblock ``Ntu rgb+d 120: A large-scale benchmark for 3d human activity
  understanding,''
\newblock {\em IEEE Transactions on Pattern Analysis and Machine Intelligence},
  vol. 42, no. 10, pp. 2684--2701, 2020.

\bibitem{RN10}
Jiang Wang, Xiaohan Nie, Yin Xia, Ying Wu, and SongChun Zhu,
\newblock ``Cross-view action modeling, learning, and recognition,''
\newblock in {\em Computer Vision and Pattern Recognition}, 2014, pp.
  2649--2656.

\bibitem{RN22}
Cong Wu, XiaoJun Wu, and Josef Kittler,
\newblock ``Graph2net: Perceptually-enriched graph learning for skeleton-based
  action recognition,''
\newblock {\em IEEE Transactions on Circuits and Systems for Video Technology},
  vol. 32, no. 4, pp. 2120--2132, 2022.

\bibitem{RN23}
Sen Yang, Xuanhan Wang, Lianli Gao, and Jingkuan Song,
\newblock ``Mke-gcn: Multi-modal knowledge embedded graph convolutional network
  for skeleton-based action recognition in the wild,''
\newblock in {\em International Conference on Multimedia and Expo (ICME)},
  2022, pp. 01--06.

\bibitem{RN27}
YuHui Wen, Lin Gao, Hongbo Fu, FangLue Zhang, Shihong Xia, and Yong-Jin Liu,
\newblock ``Motif-gcns with local and non-local temporal blocks for
  skeleton-based action recognition,''
\newblock {\em IEEE Transactions on Pattern Analysis and Machine Intelligence},
  pp. 1--1, 2022.

\bibitem{RN25}
Tianyu Guo, Hong Liu, Zhan Chen, Mengyuan Liu, Tao Wang, and Runwei Ding,
\newblock ``Contrastive learning from extremely augmented skeleton sequences
  for self-supervised action recognition,''
\newblock in {\em AAAI Conference on Artificial Intelligence}, 2022, pp.
  762--770.

\bibitem{RN26}
Kailin Xu, Fanfan Ye, Qiaoyong Zhong, and Di~Xie,
\newblock ``Topology-aware convolutional neural network for efficient
  skeleton-based action recognition,''
\newblock in {\em AAAI Conference on Artificial Intelligence}, 2022, pp.
  2866--2874.

\bibitem{RN29}
Ke~Cheng, Yifan Zhang, Congqi Cao, Lei Shi, Jian Cheng, and Hanqing Lu,
\newblock ``Decoupling gcn with dropgraph module for skeleton-based action
  recognition,''
\newblock in {\em European Conference on Computer Vision (ECCV)}, 2020, p.
  536–553.

\bibitem{RN21}
Ke~Cheng, Yifan Zhang, Xiangyu He, Jian Cheng, and Hanqing Lu,
\newblock ``Extremely lightweight skeleton-based action recognition with
  shiftgcn++,''
\newblock {\em IEEE Transactions on Image Processing}, vol. 30, pp. 7333--7348,
  2021.

\bibitem{RN31}
Kai Liu, Lei Gao, Naimul~Mefraz Khan, Lin Qi, and Ling Guan,
\newblock ``A multi-stream graph convolutional networks-hidden conditional
  random field model for skeleton-based action recognition,''
\newblock {\em IEEE Transactions on Multimedia}, vol. 23, pp. 64--76, 2021.

\bibitem{RN30}
Hao Yang, Dan Yan, Li~Zhang, Yunda Sun, Dong Li, and Stephen~J. Maybank,
\newblock ``Feedback graph convolutional network for skeleton-based action
  recognition,''
\newblock {\em IEEE Transactions on Image Processing}, vol. 31, pp. 164--175,
  2022.

\end{thebibliography}

\end{document}